\documentclass[sigconf]{acmart}
\renewcommand\footnotetextcopyrightpermission[1]{} 
\usepackage{diagbox}
\usepackage{multirow}
\usepackage{bm}
\usepackage{adjustbox}
\usepackage{xspace}
\usepackage{enumitem}

\newcommand{\bench}{\textit{MBE3.0}\xspace}
\newcommand{\model}{\textit{MOON3.0}\xspace}
\newcommand{\residual}{\textit{FIRE}\xspace}

\def\one #1 {\textbf{#1}}
\def\two #1 {\underline{#1}~}

\AtBeginDocument{%
  }

\setcopyright{acmlicensed}
\copyrightyear{2018}
\acmYear{2018}
\acmDOI{XXXXXXX.XXXXXXX}
\acmConference[Conference acronym 'XX]{Make sure to enter the correct
  conference title from your rights confirmation email}{June 03--05,
  2018}{Woodstock, NY}
\acmISBN{978-1-4503-XXXX-X/2018/06}

\acmSubmissionID{1289}

\begin{document}

\title{MOON3.0: Reasoning-aware Multimodal Representation Learning for E-commerce Product Understanding}

\author{Junxian Wu}
\authornote{Equal Contribution.}
\affiliation{
  \institution{Alibaba Group}
  \city{Hangzhou}
  \country{China}}
\email{wujunxian.wjx@taobao.com}

\author{Chenghan Fu}
\authornotemark[1]
\authornote{Project Leader.}
\affiliation{
  \institution{Alibaba Group}
  \city{Hangzhou}
  \country{China}}
\email{fuchenghan.fch@taobao.com}

\author{Zhanheng Nie}
\authornotemark[1]
\affiliation{
  \institution{Alibaba Group}
  \city{Hangzhou}
  \country{China}}
\email{niezhanheng.nzh@taobao.com}

\author{Daoze Zhang}
\authornotemark[1]
\affiliation{
  \institution{Alibaba Group}
  \city{Hangzhou}
  \country{China}}
\email{zhangdaoze.zdz@taobao.com}

\author{Bowen Wan}
\affiliation{
  \institution{Alibaba Group}
  \city{Hangzhou}
  \country{China}}
\email{wanbowen.wbw@taobao.com}

\author{Wanxian Guan}
\affiliation{
  \institution{Alibaba Group}
  \city{Hangzhou}
  \country{China}}
\email{wanxian.gwx@taobao.com}

\author{Chuan Yu}
\affiliation{
  \institution{Alibaba Group}
  \city{Hangzhou}
  \country{China}}
\email{yuchuan.yc@taobao.com}

\author{Jian Xu}
\affiliation{
  \institution{Alibaba Group}
  \city{Hangzhou}
  \country{China}}
\email{xiyu.xj@taobao.com}

\author{Bo Zheng}
\authornote{Corresponding Author.}
\affiliation{
  \institution{Alibaba Group}
  \city{Hangzhou}
  \country{China}}
\email{bozheng@alibaba-inc.com}

\renewcommand{\shortauthors}{Junxian Wu et al.}

\begin{abstract}
  With the rapid growth of e-commerce, exploring general representations rather than task-specific ones has attracted increasing attention. Although recent multimodal large language models (MLLMs) have driven significant progress in product understanding, they are typically employed as feature extractors that implicitly encode product information into global embeddings, thereby limiting their ability to capture fine-grained attributes. Therefore, we argue that leveraging the reasoning capabilities of MLLMs to explicitly model fine-grained product attributes holds significant potential. Nevertheless, achieving this goal remains non-trivial due to several key challenges: (i) long-context reasoning tends to dilute the model's attention to salient information in the raw input; (ii) supervised fine-tuning (SFT) primarily encourages rigid imitation, limiting the exploration of effective reasoning strategies; and (iii) fine-grained details are progressively attenuated during forward propagation. To address these issues, we propose \model, the first reasoning-aware MLLM-based model for product representation learning. Our method (1) employs a multi-head modality fusion module to adaptively integrate raw signals; (2) incorporates a joint contrastive and reinforcement learning framework to autonomously explore more effective reasoning strategies; and (3) introduces a fine-grained residual enhancement module to progressively preserve local details throughout the network. Additionally, we release a large-scale multimodal e-commerce benchmark \bench. Experimentally, our model demonstrates state-of-the-art zero-shot performance across various downstream tasks on both our benchmark and public datasets.

\end{abstract}

\ccsdesc[500]{Applied computing~Electronic commerce}
\ccsdesc[500]{Computing methodologies~Artificial intelligence}

\keywords{Multimodal Representations, Product Understanding, Reasoning Embedding Models, E-commerce Search}


\maketitle

\begin{figure}[t]
  \centering
  \includegraphics[width=\linewidth]{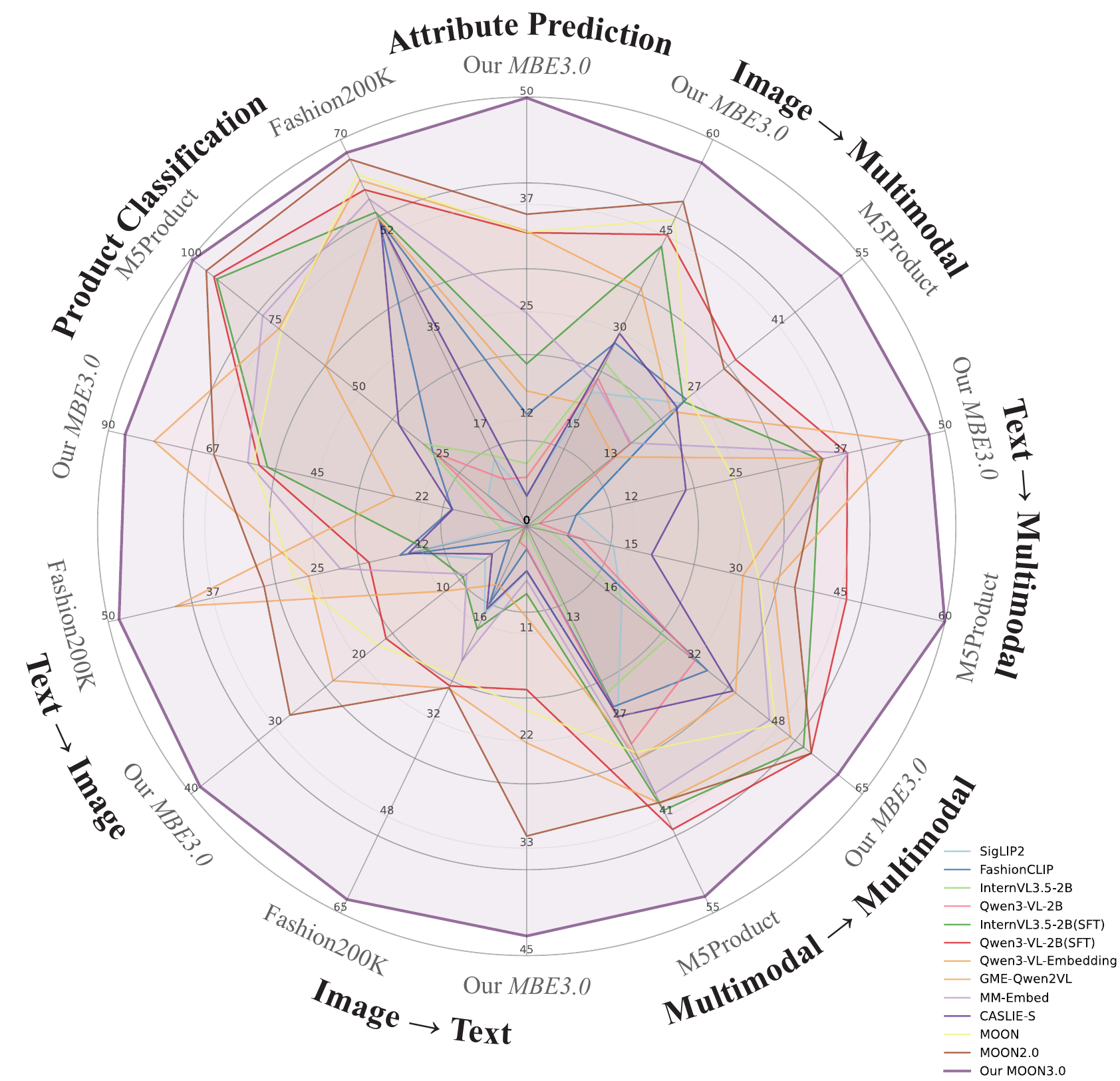}
  \caption{Overall results on all the downstream tasks.}
  \label{fig:radar}
\end{figure}

\section{Introduction}
In e-commerce scenarios, products are typically presented in multimodal formats, such as images and textual descriptions. To support downstream tasks such as product retrieval~\cite{li2021embedding,hendriksen2022multimodal,zheng2023delving,liang2025uniecs}, classification~\cite{pawlowski2022machine,xu2019open}, and attribute prediction~\cite{zhu2020multimodal}, multimodal representation learning for e-commerce aims to learn unified and discriminative embeddings for comprehensive product understanding~\cite{li2020adversarial,liu2022pretraining,yan2025mim,fu2025moon}. However, capturing fine-grained product attributes remains challenging, as it requires modeling multiple attribute factors (e.g., brand, style, and design elements) to form reliable product representations.

\begin{figure}[t]
  \centering
  \includegraphics[width=\linewidth]{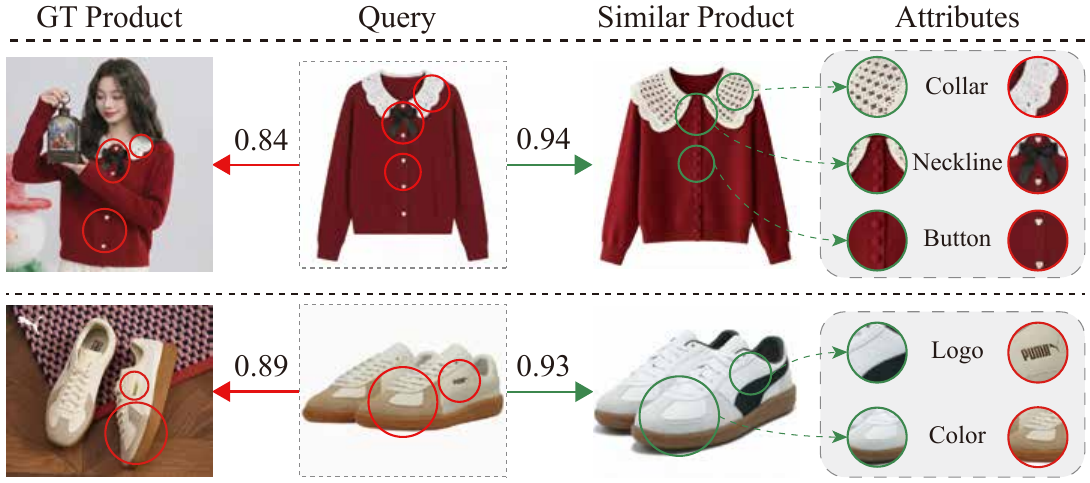}
  \caption{Comparison of fine-grained attributes among the query, the ground-truth product (denoted as ``GT Product''), and a similar product. ``$\to$'' denotes the cosine similarity between their representations.}
  \label{fig:motivation}
\end{figure}

Most existing works adopt a dual-flow architecture~\cite{yu2022commercemm,jin2023learning,wang2023missrec,jiang2024mrse,dai2024uniembedding}, which is unsuitable for modeling the many-to-one relationship in e-commerce scenarios~\cite{zhang2025moon}. Inspired by the strong multimodal understanding capabilities of MLLMs~\cite{bai2025qwen3,zhu2025internvl3}, researchers have leveraged them for representation learning~\cite{zhang2024gme,jiang2024vlm2vec,peng2024ecellm,nie2025moon2}, achieving significant improvements.
However, existing approaches treat MLLMs solely as encoders to obtain global representations, typically using the last token~\cite{zhang2024gme,jiang2024e5,lin2024mm} or mean pooling~\cite{zhang2025moon,nie2025moon2}. Under this paradigm, MLLMs are confined to feature extractors, without explicitly modeling fine-grained semantics. Consequently, models tend to capture global information, while fine-grained attributes that distinguish products are often ignored, thereby limiting the quality of the learned representations~\cite{hu2026adaptive}. 
For example, as shown in Fig.~\ref{fig:motivation}, given a red sweater with a lace collar and bowknot as the query, the model assigns a higher similarity to a visually similar product than to the ground-truth item, as dominant visual semantics such as overall garment category and color often overshadow subtle attribute differences in collar, neckline, and button design. This further reflects the underutilization of MLLMs' inherent reasoning capabilities for fine-grained product understanding.

To address these limitations and unlock the full potential of MLLMs, inspired by Chain-of-Thought (CoT)~\cite{wei2022chain} and reasoning embedding models~\cite{cui2025think,tang2025large}, we propose to utilize the reasoning ability of MLLMs to model fine-grained product attributes for multimodal representation learning in e-commerce. Despite its potential, this direction remains underexplored in current research, primarily due to the following factors.

First, long-sequence reasoning often leads to attention dilution~\cite{qin2022devil,xiao2023efficient,wu2025language,yin2024survey}. Although the autoregressively generated last token condenses semantic information, especially product attributes, the model's attention to the image-text inputs decreases as the length of the intermediate generated text increases. Consequently, the model fails to preserve subtle yet critical details from the raw inputs. Mean pooling suffers from a similar issue, as redundant tokens dilute fine-grained signals. Therefore, relying solely on the last token or mean pooling limits the representation quality for tasks requiring fine-grained understanding.

Second, SFT mainly learns to imitate predefined demonstrations, limiting the model's ability to explore more effective reasoning strategies for attribute modeling~\cite{chen2025sft, chu2025sft}. While it provides the model with preliminary reasoning and embedding capabilities, its performance is constrained by the quality of the constructed attribute annotations~\cite{liu2025uft,wang2026learning}. Specifically, the language loss in SFT relies on Teacher Forcing, enforcing strict adherence to the attribute text at each decoding step. This rigid imitation not only limits the model's inherent reasoning capabilities but also prevents the exploration of more effective attribute-level reasoning strategies.
Furthermore, most existing methods struggle to fully capture fine-grained multimodal information~\cite{jiang2024vlm2vec,gu2025breaking}, as fine-grained details are progressively attenuated during forward propagation. Recent MLLMs~\cite{bai2025qwen3,meng2024deepstack} attempt to inject fine-grained visual features into early layers of LLMs. However, this strategy remains insufficient for e-commerce representation learning. E-commerce images often contain redundant background content and unevenly informative patches, making direct injection of final-layer visual features at the LLM input introduce additional noise. Moreover, since fine-grained visual signals are fused at early layers, their influence gradually decays during forward propagation and fails to consistently support subsequent representation modeling.

To solve the challenges, we propose \model, the first reasoning-aware multimodal fine-grained product understanding model for e-commerce representation learning, which deconstructs multimodal inputs into structured, multi-dimensional attributes before generating the representation, with details as follows:

To mitigate attention dilution, we propose a multi-head modality fusion module that restores informative fine-grained signals from the raw image and text features. It uses the last token as the base representation to preserve reasoning semantics, while a gating network adaptively integrates complementary signals from the original multimodal features across distinct subspaces, resulting in more comprehensive and discriminative product representations.

To overcome the limitations of rigid imitation in SFT, we propose a joint contrastive and reinforcement learning framework for multimodal representation to further enhance the model's reasoning capabilities for attribute deconstruction. Specifically, the framework jointly optimizes embedding discriminability via a contrastive loss and reasoning quality via a Group Relative Policy Optimization (GRPO) objective with multi-dimensional composite rewards, including attribute format, information length, retrieval rankings, and attribute quality. By integrating multiple objectives, the model can explore more effective attribute reasoning strategies while maintaining discriminative and semantically consistent representations.

To address the inadequate extraction of fine-grained details, we propose a \textbf{fi}ne-grained \textbf{r}esidual \textbf{e}nhancement (\residual) module that progressively reinforces local feature modeling across visual encoding, cross-modal fusion, and language decoding stages. Instead of conventional injection, \residual preserves fine-grained signals throughout the entire forward propagation process. Specifically, during visual encoding, it applies patch-level gating to emphasize salient product regions while reducing irrelevant background interference. During cross-modal fusion, it introduces multi-level visual features into the early layers of the LLM to enhance multi-granularity semantic understanding. During language decoding, long-range residual connections re-inject shallow multimodal cues to mitigate the loss of fine-grained information in deeper layers. As a result, fine-grained details consistently contribute to representation learning, improving the discriminative power of product embeddings.

\begin{figure*}[h]
    \includegraphics[width=0.93\textwidth]{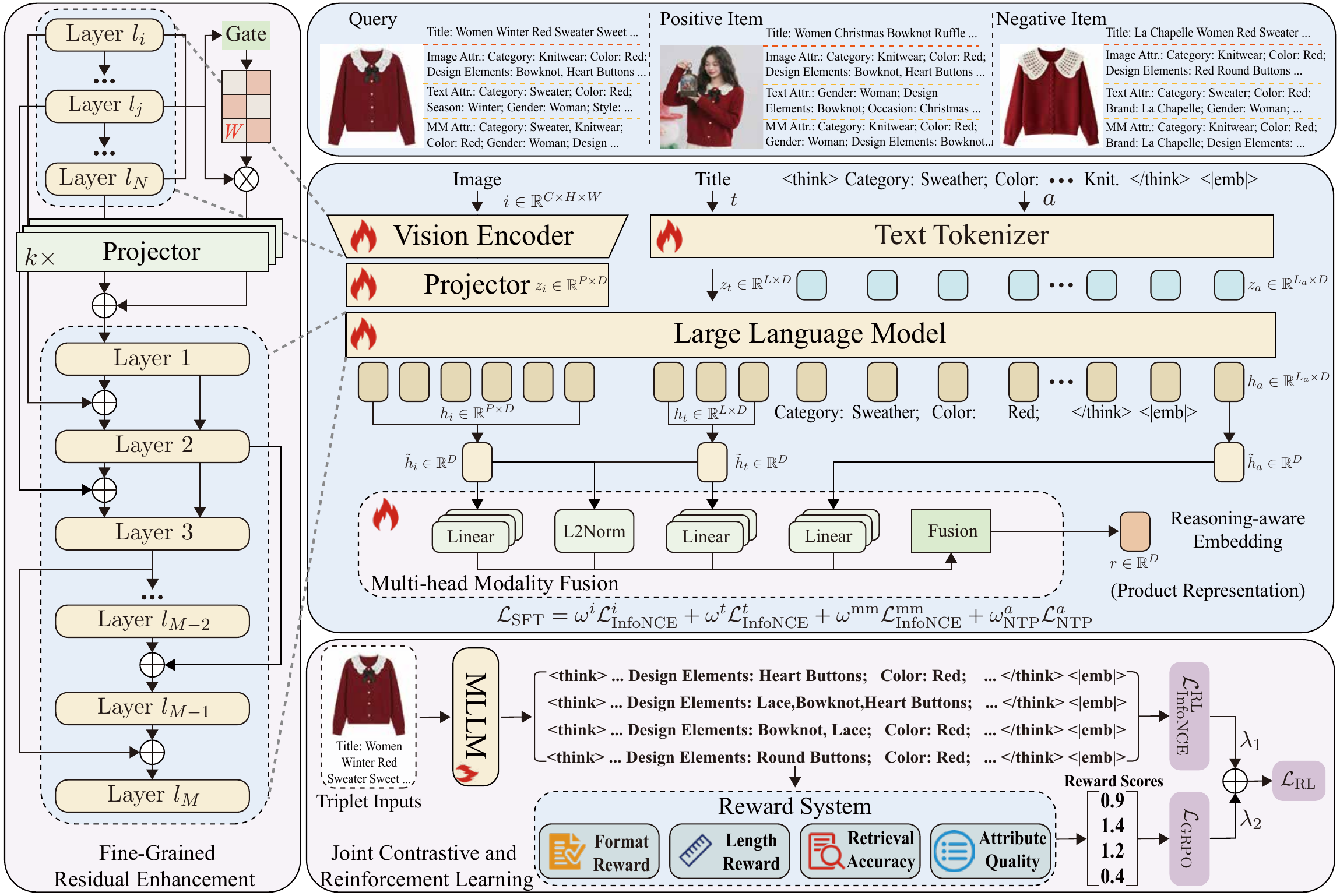}
    \caption{Pipeline of our \model. Each element in the training triplet (query, positive, and negative items) is represented in three modalities: image-only, text-only, and multimodal (image and text). ``Attr.'' denotes product attributes.}
    \label{fig:pipeline}
\end{figure*}

Furthermore, we curate a high-quality benchmark \bench for chain-of-thought attribute reasoning test by mining samples from large-scale e-commerce search logs. Extensive experiments on \bench and other public benchmarks across diverse downstream tasks demonstrate the effectiveness of our proposed model. Overall, our key contributions are summarized as follows:
\begin{itemize}[leftmargin=*]
    \item Beyond the traditional representation generation paradigm, we propose \model, the first reasoning-aware multimodal fine-grained product understanding model for e-commerce representation to enable explicit attribute deconstruction.
    \item We devise a multi-head modality fusion module and a fine-grained residual enhancement module to capture multimodal product details, and introduce a joint contrastive and reinforcement learning framework to explore reasoning strategies for fine-grained product understanding.
    \item We release a large-scale benchmark \bench, and validate \model through extensive experiments using our benchmark and other public datasets across various downstream tasks.
\end{itemize}

\section{Related Works}
\subsection{E-commerce Representation Learning}
Traditional e-commerce representation learning relies on dual-flow architectures~\cite{gao2020fashionbert,chen2022product2vec,chia2022contrastive,dai2024uniembedding,liang2025uniecs} for cross-modal alignment. However, these models struggle with the many-to-one relationships in e-commerce~\cite{zhang2025moon}.
To address this, inspired by the strong semantic capacity of MLLMs, emerging frameworks leverage MLLMs~\cite{lin2024mm,team2023gemini,zhang2024gme,achiam2023gpt,bai2023qwen} for richer product representations~\cite{ling2024captions,zhang2025moon,ling2025ecommmmu,nie2025moon2}. However, they primarily treat MLLMs as static feature extractors that directly encode multimodal inputs into embeddings, thereby limiting their ability to capture fine-grained attributes. To overcome this limitation, we propose \model, which leverages the reasoning capabilities of MLLMs to explicitly model fine-grained product attributes before generating the representation.

\subsection{MLLM Reasoning}
Chain-of-Thought (CoT) reasoning~\cite{wei2022chain,chu2024navigate} effectively mitigates visual hallucinations and improves interpretability in MLLMs~\cite{lai2024lisa,xu2025llava}. This has recently inspired a ``think-then-embed'' paradigm~\cite{yan2025o1,cui2025think} in representation learning, where models generate structured reasoning contexts before embedding them~\cite{cui2025reason,hao2026trace,zhang2025enhancing}, with recent advancements leveraging reinforcement learning to further optimize trajectory quality~\cite{jiang2026embed}.
However, reasoning-aware representation learning remains underexplored in e-commerce, where generic methods often lack domain-specific knowledge and fine-grained cross-modal alignment, highlighting the need for a dedicated multimodal representation framework within this domain. Therefore, we propose the first reasoning-aware multimodal product understanding model for e-commerce, which combines a multi-head modality fusion module, a fine-grained residual enhancement module, and a joint contrastive-reinforcement learning framework to capture intricate product nuances and enhance representation robustness.


\section{Methods}
\subsection{Problem Formulation}
In e-commerce representation learning, product understanding is evaluated through multiple multimodal downstream tasks, including multimodal retrieval, product classification, and attribute prediction~\cite{zhang2025moon,nie2025moon2}. Multimodal retrieval further covers various practical applications in e-commerce scenarios, such as image-to-multimodal, text-to-multimodal, multimodal-to-multimodal, text-to-image, and image-to-text retrieval. Together with product classification and attribute prediction, these tasks assess the model's ability to capture fine-grained cross-modal product semantics.

\textbf{Multimodal Retrieval}. Given a query $q$ and a candidate set $\mathcal{C}$, multimodal retrieval aims to identify the most relevant product. The query and candidate products, which may consist of image, text, or multimodal inputs, are mapped into a shared embedding space via an embedding function $f(\cdot)$, and retrieval is defined as:
\begin{equation}
c^* = \arg\max_{c \in \mathcal{C}} \text{sim}(f(q), f(c)).
\end{equation}

\textbf{Product Classification and Attribute Prediction}. They are formulated as embedding-based matching problems following~\cite{zhang2025moon,nie2025moon2}. Given a product $\hat{p}$, the predicted category or attribute value is defined as the candidate whose embedding is most semantically similar to the product representation:
\begin{equation}
\hat{c}^*=\arg\max_{\hat{c}\in \hat{\mathcal{C}}} \text{sim}(f(\hat{p}),f(\hat{c})),
\end{equation}
where $\hat{\mathcal{C}}$ denotes the candidate set of categories or attribute values.



\subsection{Method Overview}

As described above, diverse e-commerce tasks can be unified under an embedding-based framework for product understanding. However, most existing methods struggle to effectively capture fine-grained product attributes and fully exploit the reasoning capabilities of MLLMs. To address these limitations, we propose \model, a reasoning-aware multimodal representation model for fine-grained product understanding, which first deconstructs multimodal inputs into structured attribute-level semantics via autoregressive reasoning, and then generates unified product embeddings conditioned on these decomposed attributes. Specifically, given multimodal inputs $(i, t)$, the MLLM produces a structured rationale $a$, which represents the product as a set of multi-dimensional attribute factors: $a = \{(k_1: v_{11},v_{12},\dots), \dots, (k_N: v_{N1},v_{N2},\dots)\}$, where each $k_i$ denotes an attribute dimension (e.g., material or color) and $v_{ij}$ represents its corresponding values derived from the inputs. The rationale $a$ serves as an intermediate reasoning signal for representation generation. The preliminary representation $\tilde{h}_a$ is obtained from the hidden state of the representation token:
\begin{equation}
    \tilde{h}_a = f_\theta^{[-1]} \Big( \text{Concat}(i, t, \text{<think>}, a, \text{</think>}, \text{<|emb|>}) \Big),
\end{equation}
where $f_\theta$ denotes the MLLM forward function, and $[-1]$ indicates the hidden state of the last token.

Building upon this reasoning-aware embedding process, the overall pipeline shown in Fig.~\ref{fig:pipeline} consists of three main components. First, a multi-head modality fusion module mitigates attention dilution by using the reasoning-derived last token as the base representation and adaptively integrating complementary fine-grained signals from the original multimodal features. Second, a joint contrastive and reinforcement learning framework improves both representation discriminability and reasoning quality through contrastive objectives and GRPO-based optimization with multi-dimensional composite rewards. Third, the \residual module progressively preserves fine-grained signals across visual encoding, cross-modal fusion, and language decoding stages, enabling consistent modeling of local discriminative details throughout the forward propagation process. Together, these components produce more discriminative, interpretable, and fine-grained product representations.

\subsection{Multi-head Modality Fusion \& SFT}
\begin{figure}[t]
  \centering
  \includegraphics[width=\linewidth]{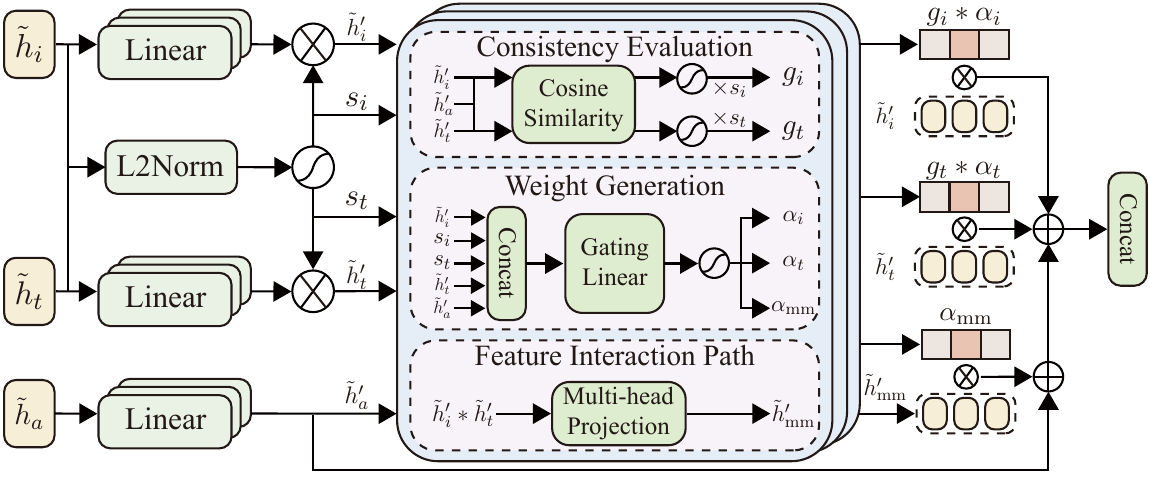}
  \caption{Multi-head Modality Fusion module.}
  \label{fig:fusion}
\end{figure}

As discussed above, relying solely on the last token or mean-pooled representations can dilute attention, because the model's focus on raw inputs gradually decays during autoregressive attribute deconstruction, causing the last token to miss crucial details. To address this issue, we introduce a multi-head modality fusion module. As shown in Fig.~\ref{fig:fusion}, the module uses the last token as the base representation to preserve reasoning semantics and leverages dynamic consistency evaluation together with multi-head subspace gating to adaptively integrate informative and complementary signals.

Specifically, we obtain raw visual and textual embeddings ($\tilde{h}_i$, $\tilde{h}_t$) by mean-pooling the MLLM hidden states of image and text tokens. Together with the last-token embedding $\tilde{h}_a$, they are fed into the module through the following pathways:

To handle inputs that may lack certain modalities (e.g., image-only or text-only queries), we compute presence scores $s_i$ and $s_t$ by applying a sigmoid function to the L2 norms of visual and textual embeddings. These scores zero out missing modalities and modulate the raw embeddings, which are linearly projected into a unified feature space to obtain $\tilde{h}'_i$ and $\tilde{h}'_t$, while the base representation $\tilde{h}_a$ is projected to $\tilde{h}'_a$. Subsequently, to ensure that only relevant signals contribute to the final representation, we dynamically evaluate semantic alignment via cosine similarity between $\tilde{h}'_a$ and each modality feature ($\tilde{h}'_i$ and $\tilde{h}'_t$), generating consistency gates $g_i$ and $g_t$ for subsequent fusion steps.


Having ensured that only the relevant and explicitly present signals are retained, we further address the attenuation of fine-grained details during the reasoning process. Inspired by the multi-head attention mechanism~\cite{vaswani2017attention}, instead of applying a coarse global scalar weight to the entire representation, we partition the features into $H$ distinct semantic subspaces. To achieve adaptive integration within each subspace, the projected raw features, base representation, and presence scores $(\tilde{h}'_i, \tilde{h}'_t, \tilde{h}'_a, s_i, s_t)$ are concatenated and fed into a linear gating network that models cross-modal feature importance, producing head-specific fusion weights $\alpha_i$, $\alpha_t$, and $\alpha_{\text{mm}}$. This design enables the network to selectively restore the most relevant visual and textual details within each semantic subspace.

Beyond restoring individual modalities, the autoregressive process may cause the last token to lose shared image–text semantics. To recover this joint information, we compute the Hadamard product $\tilde{h}'_{\text{mm}}=\tilde{h}'_i \odot \tilde{h}'_t$, which captures the element-wise co-activation and intersection of both modalities. This interactive feature is modulated by a joint weight $\alpha_{\text{mm}}$ to reinforce shared multimodal signals.

Ultimately, the final representation $r$ is obtained by fusing the base representation with the adaptively weighted individual and interactive signals:
\begin{equation}
    r=\tilde{h}'_a+(g_i \cdot \alpha_i) \cdot \tilde{h}'_i+(g_t \cdot \alpha_t) \cdot \tilde{h}'_t + \alpha_{\text{mm}} \cdot \tilde{h}'_{\text{mm}}.
\end{equation}

This fusion mitigates attention dilution, ensuring that the multi-head representation $r$ preserves both explicit reasoning semantics and fine-grained multimodal input features.

\par After obtaining the final representation $r$, we proceed to optimize the model. To enable the MLLM to autoregressively reason in the desired attribute-deconstruction format while generating discriminative embeddings for downstream tasks, we adopt a joint optimization strategy with a cross-entropy loss for next-token prediction and an InfoNCE loss for contrastive learning during SFT.

\par Specifically, to encourage the model to follow a consistent structured reasoning process across diverse settings, our generation objective considers multiple input configurations. For each training instance $n$, let $\mathcal{S}_n=\{q_i,q_t,q_{\text{mm}},p_{\text{mm}},n_{\text{mm}}\}$ denote the set of corresponding diverse inputs. For each input $x\in \mathcal{S}_n$, the model generates a structured reasoning sequence $Y^x=(y_1^x,y_2^x,...,y_{L_x}^x)$. The unified generation loss $\mathcal{L}_\text{NTP}^a$ aggregates the cross-entropy losses over all inputs:
\begin{equation}
\mathcal{L}_{\text{NTP}}^a = -\frac{1}{\mathcal{B}} \sum_{i=1}^{\mathcal{B}} \sum_{x \in \mathcal{S}_i} \sum_{j=1}^{L_x} \log P(y_{j}^x | x, y_{<j}^x),
\end{equation}
where $\mathcal{B}$ is the mini-batch size, $L_x$ is the reasoning length for input $x$, and $y_j^x$ is the $j$-th generated token. 

Furthermore, we adopt the InfoNCE loss in a contrastive learning framework to learn discriminative representations. We construct training triplets $(q,p,n)$, denoting a query, a positive, and a negative. Let $r_q$, $r_p$, and $r_n$ denote their corresponding representations. The InfoNCE loss $\mathcal{L}_{\text{InfoNCE}}$ is defined as:
\begin{equation}
\mathcal{L}_{\text{InfoNCE}} = - \frac{1}{\mathcal{B}} \sum_{i=1}^{\mathcal{B}} \log \frac{\exp(r_q \cdot r_p / \tau)}{\exp(r_q \cdot r_p / \tau) + \sum_{\mathcal{N}_q} \exp(r_q \cdot r_n / \tau)},
\end{equation}
where $\tau$ is the temperature parameter, and $\mathcal{N}_q$ denotes a set of negative items including in-batch and hard samples.

To support diverse e-commerce search scenarios, the query $q$ is considered under three modality settings: image-only ($i$), text-only ($t$), and multimodal ($\text{mm}$). Accordingly, we compute modality-specific InfoNCE losses, denoted as $\mathcal{L}^{i}_{\text{InfoNCE}}$, $\mathcal{L}^{t}_{\text{InfoNCE}}$, and $\mathcal{L}^{\text{mm}}_{\text{InfoNCE}}$, to align each query modality with its corresponding multimodal item representation.

Finally, the training objective is defined as a weighted sum of the modality-specific contrastive losses and the NTP loss:
\begin{equation}
\mathcal{L}_\text{SFT}=\omega^{i}\mathcal{L}^{i}_{\text{InfoNCE}}+\omega^{t}\mathcal{L}^{t}_{\text{InfoNCE}}+\omega^{\text{mm}}\mathcal{L}^{\text{mm}}_{\text{InfoNCE}}+\omega_{\text{NTP}}^a\mathcal{L}_{\text{NTP}}^a,
\end{equation}
where $\omega^{i}$, $\omega^{t}$, and $\omega^{\text{mm}}$ are modality-specific coefficients controlling the relative importance of each query type, and $\omega_{\text{NTP}}^a$ is a coefficient balancing the contributions of representation learning and generative reasoning.

\subsection{Joint Contrastive \& Reinforcement Learning}
To overcome the limitations of rigid imitation inherent in SFT and enable the model to explore more effective reasoning strategies, we introduce a joint contrastive and reinforcement learning framework. Under the structured reasoning formulation, the framework jointly optimizes both reasoning quality and representation discriminability. Specifically, given an arbitrary element of a triplet as input, the MLLM acts as a policy model $\pi_\theta$ and samples a group of $G$ reasoning trajectories, denoted as $\{a_1,a_2,...,a_G\}$. Each trajectory autoregressively generates a structured attribute sequence along with a candidate representation $r_g$.

To evaluate the quality of each sampled trajectory, we design a reward system that assigns a composite score $u_g$:
\begin{equation}
u_g = \omega_1 u_{\text{format}} + \omega_2 u_{\text{length}} + \omega_3 u_{\text{accuracy}} + \omega_4 u_{\text{quality}},
\end{equation}
where $\omega_1$, $\omega_2$, $\omega_3$, and $\omega_4$ are the reward coefficients. The individual reward components are defined as follows:

\textbf{Format and Length Rewards}. To enforce proper output format, $u_{\text{format}}$ evaluates structural compliance, assigning 1 if the trajectory follows the correct sequential order of the predefined structural and embedding states, and 0 otherwise. To prevent excessive verbosity and latency, $u_{\text{length}}$ yields 1 if the token length of $a_g$ is within a target threshold, and 0 if it exceeds the limit.

\textbf{Retrieval Accuracy Reward}. To ensure that the representation $r_g$ can successfully retrieve the positive item $p_g$, we compute the cosine similarity between $r_g$ and a global item pool $\mathcal{N}_{\text{pool}}$, including all positive and negative items across devices. Let rank($p_g$) denote the descending similarity rank of the positive item $p_g$ within this pool, with 1 indicating the highest similarity. We then define a continuous logarithmic rank reward $u_{\text{accuracy}}$ as:
\begin{equation}
u_{\text{accuracy}} = 1.0 - \log(\text{rank}(p_g)) /
\log(|\mathcal{N}_{\text{pool}}|).
\end{equation}
This reward assigns 1.0 for top-1 retrieval and decays smoothly toward 0.0 for lower ranks, providing dense signals for exploration.

\begin{figure*}[!t]
  \centering
  \includegraphics[width=\textwidth]{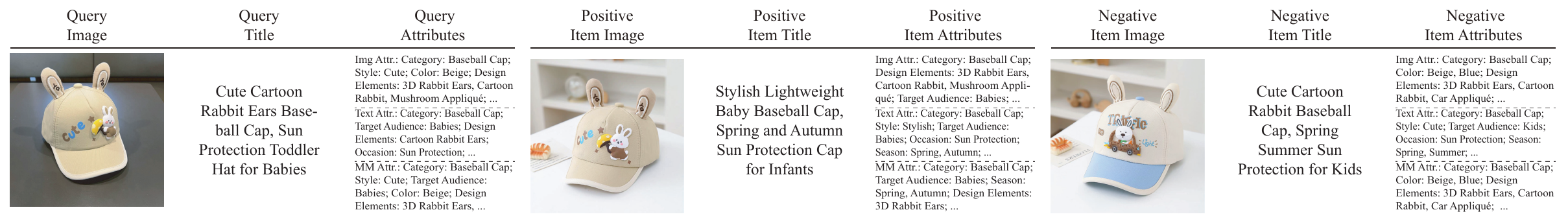}
  \caption{A training sample of our \bench benchmark. Each training sample includes the image and title of the query, positive item, and negative item, as well as three types of attributes: image-derived attributes (“Img Attr.”), text-derived attributes (“Text Attr.”), and multimodal attributes (“MM Attr.”, the union of image and text attributes).}
  \label{fig:benchmark}
\end{figure*}

\textbf{Attribute Quality Reward}. To encourage the model to generate informative attribute key-value pairs, we define a reward $u_{\text{quality}}$ that evaluates both semantic richness and factual consistency with the original label. Factuality is assessed using a distilled MLLM-based evaluator, which takes the label, input image, and generated attributes as inputs, and outputs a normalized score $s_{\text{MLLM}} \in [0,1]$ that quantifies the information gain provided by the generated attributes while ensuring consistency with the original inputs. Meanwhile, to balance informativeness and conciseness, extra attribute values beyond the label are rewarded up to a threshold $\tau_q$. Let $N_g$ and $N_l$ denote the total number of generated and label key-value pairs, respectively. The attribute quality reward is defined as:
\begin{equation}
u_{\text{quality}} = s_{\text{MLLM}} + \alpha_q \cdot \min(\max(0, N_g - N_l), \tau_q),
\end{equation}
where $\alpha_q$ is a scaling factor and $\tau_q$ is the maximum useful number of additional attributes. This formulation encourages generating more key-value pairs than the label while mitigating excessive verbosity.

Based on the evaluated rewards, we employ the GRPO algorithm to refine the reasoning policy without an external value model. The advantage $A_g$ for each trajectory is computed by normalizing the rewards: $A_g=(u_g-\mu)/\sigma$, where $\mu$ and $\sigma$ are the mean and standard deviation of $\{u_1,u_2,...,u_G\}$. The policy is then optimized using:
\begin{equation}
\mathcal{L}_{\text{GRPO}} = - \frac{1}{G} \sum_{g=1}^{G} \min \left( \rho_g A_g, \text{clip}(\rho_g, 1 - \epsilon, 1 + \epsilon) A_g \right),
\end{equation}
where $\rho_g=\frac{\pi_\theta(a_g|q)}{\pi_{\text{old}}(a_g|q)}$ is the probability ratio between the active policy and the old policy, and $\epsilon$ is the clipping threshold.

To guarantee that the dynamically explored representations remain highly discriminative, we maintain the modality-specific InfoNCE contrastive constraints across the $G$ generations. Ultimately, the joint learning phase fuses these complementary objectives:
\begin{equation}
\mathcal{L}_{\text{RL}} = \lambda_1 \mathcal{L}_{\text{InfoNCE}}^{\text{RL}} + \lambda_2 \mathcal{L}_{\text{GRPO}},
\end{equation}
where $\lambda_1$ and $\lambda_2$ are the loss coefficients.

By integrating these objectives, our framework encourages the model to discover superior reasoning strategies for attribute deconstruction while preserving representation quality.

\subsection{Fine-Grained Residual Enhancement}
Fine-grained multimodal details are essential for accurate product representation learning. However, conventional feature injection in MLLMs introduces background noise in e-commerce images and causes semantic attenuation during deep forward propagation. To mitigate these issues, we propose the fine-grained residual enhancement (\residual) module, which progressively reinforces local feature modeling to preserve fine-grained details across three key stages:

\textbf{Visual Encoding Stage}. E-commerce images often contain background noise, and the relevance of patches varies across regions. To emphasize main product details, we introduce patch-level gated residual connections. Let $X^L \in \mathbb{R}^{P \times D}$ be the final-layer patch embedding from the vision encoder, and $\{X^m\}_{m=1}^{V}$ the intermediate patch features from earlier layers. To adaptively compute the relative contribution of each patch, we calculate a spatial gating coefficient using both intermediate and final semantics: $\lambda^m = \sigma \big( f_m([X^m; X^L]) \big)$, where $f_m(\cdot)$ is a lightweight MLP, $[\cdot;\cdot]$ denotes feature concatenation, and $\sigma$ is the sigmoid activation. These adaptively filtered patch features are then injected into the final-layer embedding:
\begin{equation}
X^{\text{enh}} = \text{LayerNorm} \Big( X^L + \sum_{m=1}^{V} \lambda^m X^m \Big).
\end{equation}

\textbf{Cross-modal Fusion Stage}. To overcome the limited multi-granularity visual information fusion, and inspired by~\cite{bai2025qwen3}, \residual extracts visual features from a selected set of intermediate vision layers $L^m=\{l_1^m,...,l_n^m\}$. Each multi-scale feature is independently projected via a cross-modal projector and injected into multiple early layers of the LLM. This explicit injection provides the model with a hierarchical visual representation encompassing fine-grained local details, specific design patterns, and holistic product semantics, thereby facilitating the subsequent attribute deconstruction.

\textbf{Language Decoding Stage}. Fine-grained multimodal cues can diminish in the deeper layers of the LLM during forward propagation. To address this, \residual introduces long-range residual connections. Let $h_{\text{early}}$ denote the multimodal hidden state after initial fusion, and $h_{\text{deep}}$ denote a deeper state before the last layer. We explicitly re-inject the shallow multimodal cues into $h_{\text{deep}}$ as follows:
\begin{equation}
h_{\text{deep}}' = h_{\text{deep}} + \alpha \cdot W_r (h_{\text{early}}),
\end{equation}
where $\alpha$ denotes a learnable scaling parameter, and $W_r$ represents a linear projection weight.

Through this holistic design, the hierarchically preserved fine-grained details continuously contribute to the representation, enhancing the model's ability to capture nuanced information.

\begin{table*}[!t]
    \caption{Zero-shot results of the multimodal retrieval, product classification, and attribute prediction tasks on our \bench benchmark. Both Qwen3-VL-Embedding and GME-Qwen2VL are based on 2B-parameter models. ``Dim.'' denotes the dimension of representations output by the models (all correspond to the models' original output).}
    \label{tab:ourbench}
    \centering
    \setlength\tabcolsep{3pt}
    \renewcommand{\arraystretch}{1.0}  
\begin{adjustbox}{width=\textwidth,center}
\begin{tabular}{lcrrrrrrrrrrrrrrrrrrrrrrr}
\toprule
\multicolumn{1}{c}{\multirow{3}{*}{\diagbox{Methods}{Metrics}}} & 
\multicolumn{1}{c}{\multirow{3}{*}{Dim.}} &
\multicolumn{15}{c}{Multimodal Retrieval}& 
\multicolumn{4}{c}{\multirow{2}{*}{Product Classification}} &
\multicolumn{4}{c}{\multirow{2}{*}{Attribute Prediction}} \\
\cmidrule(lr){3-17}

&
\multicolumn{1}{c}{} &
\multicolumn{3}{c}{$q^{i} \to c^{\text{mm}}$} & 
\multicolumn{3}{c}{$q^{t} \to c^{\text{mm}}$} & 
\multicolumn{3}{c}{$q^{\text{mm}} \to c^{\text{mm}}$} &
\multicolumn{3}{c}{$q^{i} \to c^{t}$} &
\multicolumn{3}{c}{$q^{t} \to c^{i}$} \\
\cmidrule(lr){3-5} \cmidrule(lr){6-8} \cmidrule(lr){9-11} \cmidrule(lr){12-14} \cmidrule(lr){15-17} \cmidrule(lr){18-21} \cmidrule(lr){22-25}
                  & \multicolumn{1}{c}{} 
                  & \multicolumn{1}{c}{R@1} & \multicolumn{1}{c}{R@5} & \multicolumn{1}{c}{R@10} & \multicolumn{1}{c}{R@1} & \multicolumn{1}{c}{R@5} & \multicolumn{1}{c}{R@10} & \multicolumn{1}{c}{R@1} & \multicolumn{1}{c}{R@5} & \multicolumn{1}{c}{R@10} & \multicolumn{1}{c}{R@1} & \multicolumn{1}{c}{R@5} & \multicolumn{1}{c}{R@10} & \multicolumn{1}{c}{R@1} & \multicolumn{1}{c}{R@5} & \multicolumn{1}{c}{R@10} &
                  \multicolumn{1}{c}{Acc.} & \multicolumn{1}{c}{Prec.} & \multicolumn{1}{c}{Rec.} & \multicolumn{1}{c}{F1} &
                  \multicolumn{1}{c}{Acc.} & \multicolumn{1}{c}{Prec.} & \multicolumn{1}{c}{Rec.} & \multicolumn{1}{c}{F1}\\
\midrule
SigLIP2~\cite{tschannen2025siglip}
& 3072
& 5.31 & 14.95 & 20.83 
& 2.07 & 4.46 & 6.06 
& 5.55 & 13.61 & 18.42 
& 1.10 & 3.13 & 4.78 
& 0.79 & 3.02 & 4.95 
& 1.47 & 1.88 & 1.83 & 0.53 
& 3.66 & 2.21 & 4.98 & 1.06 \\									
FashionCLIP~\cite{chia2022contrastive} 
& 512
& 8.27 & 22.52 & 28.44
& 1.49 & 4.33 & 5.91
& 10.15 & 28.31 & 35.01 
& 0.53 & 1.61 & 2.37 
& 0.47 & 1.42 & 2.06 
& 16.08 & 19.68 & 14.03 & 11.40 
& 13.05 & 15.44 & 12.93 & 9.44 \\
\midrule
InternVL3.5-2B~\cite{wang2025internvl3}  
& 2048
& 8.01 & 20.69 & 25.39 
& 0.09 & 0.35 & 0.60 
& 8.26 & 22.14 & 27.24 
& 0.09 & 0.36 & 0.65 
& 0.05 & 0.19 & 0.34 
& 7.84 & 16.61 & 7.87 & 4.78 
& 7.32 & 11.12 & 9.528 & 4.53 \\	
Qwen3-VL-2B~\cite{bai2025qwen3}  
& 2048
& 6.75 & 18.09 & 22.96 
& 0.26 & 0.94 & 1.55 
& 9.37 & 26.03 & 32.58 
& 0.34 & 1.35 & 2.27 
& 0.04 & 0.18 & 0.32 
& 4.83 & 8.55 & 4.33 & 1.70 
& 5.72 & 10.07 & 7.57 & 3.10 \\	
InternVL3.5-2B (SFT)~\cite{wang2025internvl3}  
& 256
& 13.04 & 34.92 & 43.44 
& 8.51 & 25.11 & 35.05 
& 15.22 & 43.17 & 53.64 
& 1.58 & 4.89 & 7.07 
& 1.85 & 5.33 & 7.58 
& 55.77 & 52.54 & 61.08 & 52.77 
& 18.93 & 18.05 & 21.86 & 16.98 \\
Qwen3-VL-2B (SFT)~\cite{bai2025qwen3}  
& 256
& 13.21 & 35.37 & 45.26 
& 9.00 & 27.12 & 38.33 
& 15.51 & 44.53 & 55.09 
& 3.31 & 10.06 & 17.12 
& 3.01 & 11.13 & 16.77 
& 57.52 & 59.13 & 69.34 & 59.84 
& 34.21 & 35.59 & 40.33 & 32.5 \\
\midrule	
Qwen3-VL-Embedding~\cite{li2026qwen3} 
& 2048
& 5.06 & 14.42 & 18.94 
& 9.97 & 27.80 & 35.56
& 11.27 & 32.60 & 40.47 
& 2.35 & 6.83 & 9.56 
& 2.65 & 7.10 & 9.68 
& 28.46 & 58.87 & 25.67 & 24.66 
& 15.78 & 38.15 & 18.89 & 13.95 \\
GME-Qwen2VL~\cite{zhang2024gme} 
& 1536
& 9.58 & 27.84 & 36.91 
& \two{11.33} & \two{34.49} & \two{44.90}
& 13.97 & 40.69 & 51.12 
& 5.13 & 15.84 & 22.71 
& \two{5.78} & 16.52 & 23.08 
& \two{80.17} & \two{78.51} & \two{79.05} & \two{76.17} 
& 34.41 & \two{41.39} & \two{41.47} & 33.62 \\
MM-Embed~\cite{lin2024mm}  
& 4096
& 6.06	& 16.70	& 22.30 
& 10.04	& 29.31	& 38.41 
& 12.88	& 37.45	& 47.04 
& 1.01	& 3.56	& 5.66 
& 1.40	& 4.62	& 7.15 
& 59.99	& 63.36	& 56.58	& 50.89 
& 24.85	& 30.95	& 24.13	& 18.49 \\
\midrule
CASLIE-S~\cite{ling2024captions} 
& 3072
& 9.02	& 24.15	& 29.94 
& 5.61	& 14.28	& 19.02 
& 12.68	& 32.58	& 39.99 
& 1.00	& 2.86	& 4.67 
& 0.76	& 2.65	& 4.12 
& 15.93	& 27.88	& 11.12	& 7.99
& 3.54	& 3.70	& 4.11	& 1.77\\	
MOON~\cite{zhang2025moon} 
& 256
& 13.55	& \two{40.61}	& 47.65
& 5.89	& 17.69	& 24.76 
& 14.35	& 40.03	& 48.20
& 4.51	& 13.62	& 19.30
& 4.11	& 12.44	& 17.76
& 58.79	& 55.16	& 62.36	& 56.53
& 34.28	& 30.21	& 37.00	& 31.39 \\
MOON2.0~\cite{nie2025moon2}
& 256
& \two{14.08}	& 40.19	& \two{50.41}
& 8.73	& 26.01	& 35.33
& \two{15.65}	& \two{45.05}	& \two{55.15}
& \two{6.62}	& \two{21.90}	& \two{32.47}
& 5.51	& \two{18.63}	& \two{28.22}
& 67.27	& 63.26	& 72.72	& 65.46 
& \two{36.36}	& 39.56	& 40.48	& \two{35.11} \\
\midrule					
Our \model 
& 256
& \one{16.14}	& \one{44.68}	& \one{56.39}
& \one{12.57}	& \one{36.59}	& \one{48.10}
& \one{16.79}	& \one{48.11}	& \one{60.30}
& \one{11.39}	& \one{31.98}	& \one{42.95}
& \one{10.24}	& \one{28.68}	& \one{38.93}
& \one{86.40}	& \one{84.43}	& \one{86.79}	& \one{83.98}
& \one{49.92}	& \one{51.67}	& \one{57.54}	& \one{49.48} \\
\bottomrule
\end{tabular}
\end{adjustbox}
\end{table*}

\begin{table*}[!t]
    \caption{Zero-shot results of the multimodal retrieval and product classification tasks on the M5Product and Fashion200K benchmarks. Both Qwen3-VL-Embedding and GME-Qwen2VL are based on 2B-parameter models.}
    \label{tab:m5product}
    \centering
    \setlength\tabcolsep{3pt}
    \renewcommand{\arraystretch}{1.0}  
\begin{adjustbox}{width=\textwidth,center}
\begin{tabular}{lrrrrrrrrrrrrrrrrrrrrrrr}
\toprule
\multicolumn{1}{c}{\multirow{4}{*}{\diagbox{Methods}{Metrics}}} & 
\multicolumn{15}{c}{Multimodal Retrieval}& 
\multicolumn{8}{c}{Product Classification} \\
\cmidrule(lr){2-16} \cmidrule(lr){17-24}
&
\multicolumn{9}{c}{M5Product}&
\multicolumn{6}{c}{Fashion200K}&
\multicolumn{4}{c}{\multirow{2}{*}{M5Product}}&
\multicolumn{4}{c}{\multirow{2}{*}{Fashion200K}} \\
\cmidrule(lr){2-10} \cmidrule(lr){11-16}
&
\multicolumn{3}{c}{$q^{i} \to c^{\text{mm}}$} & 
\multicolumn{3}{c}{$q^{t} \to c^{\text{mm}}$} & 
\multicolumn{3}{c}{$q^{\text{mm}} \to c^{\text{mm}}$} &
\multicolumn{3}{c}{$q^{i} \to c^{t}$} &
\multicolumn{3}{c}{$q^{t} \to c^{i}$} \\
\cmidrule(lr){2-4} \cmidrule(lr){5-7} \cmidrule(lr){8-10} \cmidrule(lr){11-13} \cmidrule(lr){14-16} \cmidrule(lr){17-20} \cmidrule(lr){21-24}
                  & \multicolumn{1}{c}{R@1} & \multicolumn{1}{c}{R@5} & \multicolumn{1}{c}{R@10} & \multicolumn{1}{c}{R@1} & \multicolumn{1}{c}{R@5} & \multicolumn{1}{c}{R@10} & \multicolumn{1}{c}{R@1} & \multicolumn{1}{c}{R@5} & \multicolumn{1}{c}{R@10} & \multicolumn{1}{c}{R@1} & \multicolumn{1}{c}{R@5} & \multicolumn{1}{c}{R@10} & \multicolumn{1}{c}{R@1} & \multicolumn{1}{c}{R@5} & \multicolumn{1}{c}{R@10} &
                  \multicolumn{1}{c}{Acc.} & \multicolumn{1}{c}{Prec.} & \multicolumn{1}{c}{Rec.} & \multicolumn{1}{c}{F1} &
                  \multicolumn{1}{c}{Acc.} & \multicolumn{1}{c}{Prec.} & \multicolumn{1}{c}{Rec.} & \multicolumn{1}{c}{F1}\\
\midrule				
SigLIP2~\cite{tschannen2025siglip} 
& 8.09	& 19.48	& 25.29 
& 2.41	& 8.49	& 12.44 
& 9.53	& 21.71	& 26.87 
& 4.39	& 10.49	& 14.70 
& 4.19	& 10.03	& 13.90 
& 11.46	& 24.13	& 12.14	& 9.71 
& 12.18	& 20.04	& 10.92	& 6.6 \\
FashionCLIP~\cite{chia2022contrastive}
& 8.23	& 19.44	& 26.13
& 1.16	& 2.53	& 5.88 
& 9.21	& 20.76	& 25.65
& 4.34	& 9.15	& 12.97
& 4.92	& 11.67	& 15.14
& 28.02	& 44.82	& 26.8	& 22.39
& 55.42	& 65.39	& 63.13	& 54.66 \\
\midrule
InternVL3.5-2B~\cite{wang2025internvl3}  
& 5.83	& 15.16	& 21.02
& 0.68	& 2.31	& 3.68 
& 6.67	& 17.26	& 23.70
& 0.58	& 1.96	& 3.14 
& 0.36	& 1.49	& 2.60 
& 30.81	& 48.03	& 29.58	& 27.51
& 12.83	& 22.82	& 18.14	& 18.34 \\	
Qwen3-VL-2B~\cite{bai2025qwen3}  
& 5.38	& 13.1	& 16.91 
& 1.88	& 5.38	& 7.89 
& 9.14	& 20.07	& 30.95
& 0.49	& 1.59	& 2.57
& 0.08	& 0.19	& 0.30
& 27.51	& 32.82	& 27.07	& 21.53
& 8.52	& 17.66	& 14.83	& 12.87 \\	
InternVL3.5-2B (SFT)~\cite{wang2025internvl3}  
& 7.92	& 19.31	& 25.66 
& 13.51	& 32.38	& 41.13 
& 13.67	& 27.40	& 40.40
& 5.04	& 12.39	& 17.27 
& 3.30	& 8.37	& 11.97
& 92.38	& 92.09	& 92.72	& 92.08
& 56.84	& 58.2	& 60.65	& 52.67 \\
Qwen3-VL-2B (SFT)~\cite{bai2025qwen3}  
& 10.60	& 23.94	& \two{34.25} 
& \two{15.79}	& \two{33.17}	& \two{45.87}
& 14.53	& \two{29.23}	& \two{43.11}
& 10.68	& 22.79	& 26.83
& 5.16	& 13.30	& 18.85
& 93.23	& 93.23	& 93.22	& 93.21
& 60.92	& 65.50	& 67.83	& 61.28 \\
\midrule	
Qwen3-VL-Embedding~\cite{li2026qwen3} 
& 4.51	& 10.67	& 14.25
& 10.43	& 24.36	& 31.00
& 11.81	& 26.55	& 32.98
& 2.56	& 6.65	& 9.76
& \two{14.14}	& \two{31.87}	& \two{41.98}
& 59.98	& 91.59	& 59.22	& 60.83 
& 55.76	& 73.36	& 57.55	& 54.91 \\
GME-Qwen2VL~\cite{zhang2024gme}
& 6.66	& 17.87	& 23.60
& 13.96	& 24.25	& 35.42
& 14.49	& 26.61	& 39.40
& 10.31	& 21.14	& \two{27.23}
& 8.13	& 20.31	& 26.06
& 73.87	& 73.68	& 74.06	& 73.4
& 62.66	& 63.44	& 68.15	& 61.38 \\
MM-Embed~\cite{lin2024mm}
& 5.78 & 13.57 & 17.14 
& 11.90 & 22.36 & 33.23 
& 13.43 & 25.31 & 37.96
& 6.12 & 15.91 & 22.57 
& 5.97 & 15.71 & 22.25 
& 78.70	& 83.71	& 77.91	& 76.05 
& 59.24 & 66.29 & 66.60 & 59.24 \\
\midrule
CASLIE-S~\cite{ling2024captions} 
& 8.00 & 19.50 & 24.57 
& 5.69 & 11.13 & 17.90 
& 8.40 & 20.44 & 27.04
& 4.41 & 10.04 & 13.89 
& 4.71 & 11.25 & 14.12
& 38.16	& 57.08	& 36.64	& 30.39
& 54.88 & 58.23 & 55.65 & 53.76\\	
MOON~\cite{zhang2025moon}
& 9.11 & 21.59 & 26.53
& 10.14 & 20.33 & 33.31
& 14.27 & 25.05 & 32.13
& 11.71 & 20.02 & 25.09
& 10.82 & 22.89 & 27.73
& 73.12	& 66.23	& 70.00	& 66.43
& 63.74	& 59.02	& 68.00	& 61.05 \\
MOON2.0~\cite{nie2025moon2}
& \two{11.28}	& \two{24.43}	& 32.37
& 15.27	& 25.69	& 38.45
& \two{15.21}	& 27.37	& 39.27
& \two{13.1}	& \two{23.16}	& 27.09
& 13.05	& 25.25	& 31.39
& \two{95.50}	& \two{95.42}	& \two{95.30}	& \two{95.18} 
& \two{66.44}	& \two{68.9}	& \two{69.55}	& \two{64.21} \\
\midrule					
Our \model 
& \one{15.78}	& \one{39.78}	& \one{51.49}
& \one{18.27}	& \one{46.84}	& \one{59.96}
& \one{16.51}	& \one{41.08}	& \one{52.64}
& \one{27.10}	& \one{51.96}	& \one{62.72}
& \one{18.53}	& \one{38.98}	& \one{48.72}
& \one{99.54}	& \one{99.57}	& \one{99.51}	& \one{99.54}
& \one{67.65}	& \one{77.62}	& \one{74.05}	& \one{66.99} \\
\bottomrule
\end{tabular}
\end{adjustbox}
\end{table*}

\section{The \bench Benchmark}
Training reasoning-aware representation models relies on high-quality CoT supervision to capture structured attributes from product inputs. However, such annotations are rarely available in existing e-commerce datasets. To address this, we construct a large-scale reasoning-aware multimodal benchmark \bench, derived from user behavior logs on a Chinese e-commerce platform.

Specifically, we form query-positive-negative triplets $(q,p,n)$ with purchased items as positives and exposed-but-unclicked items as negatives. For reasoning supervision, we build a large-scale multimodal CoT corpus via a multi-stage annotation pipeline.

For product images, we employ a strong MLLM (e.g., Qwen3-VL-32B-Instruct) to generate fine-grained captions, which are then converted into structured attribute sequences (e.g., category, material, and design elements). For product titles, the model directly generates attribute-level reasoning sequences in the same space.
To ensure annotation reliability, we adopt a multimodal relevance model to verify consistency between the generated reasoning and the original product inputs to remove hallucinated annotations.


Following this pipeline, the training set contains \textbf{7,691,199} samples with reasoning supervision.
The test set was further annotated with positive item categories and subjected to rigorous manual verification for attribute and label accuracy, yielding \textbf{916,188} high-quality samples for retrieval, classification, and attribute prediction tasks. To make the benchmark data more intuitive, an example of the constructed training samples is illustrated in Fig.~\ref{fig:benchmark}.

\section{Experiments}
\subsection{Experimental Setup}
\textbf{Training}. We initialize the model from Qwen3-VL-2B-Instruct~\cite{bai2025qwen3} and train it with SFT followed by joint contrastive and reinforcement learning on our proposed training dataset. For SFT, we use a per-GPU batch size of 16 and a learning rate of $5\times10^{-5}$ with a cosine scheduler. The model is trained on 64 GPUs (NVIDIA H20) for approximately 38 hours. The loss coefficients are set to $\omega^{i}=1$, $\omega^{t}=0.3$, $\omega^{\text{mm}}=0.1$, and $\omega_{\text{NTP}}^a=0.01$. During reinforcement learning, we set the group size $G=8$, with reward weights $\omega_{1}=0.5$, $\omega_{2}=0.3$, $\omega_{3}=1$, and $\omega_{4}=1$, and loss weights $\lambda_1=0.1$ and $\lambda_2=1$. In the attribute quality reward, $\alpha_q=0.2$ and $\tau_q=4$. We use a per-GPU batch size of 8 and a learning rate of $5\times10^{-6}$. This stage takes approximately 12 hours on 64 GPUs (NVIDIA H20).

\textbf{Evaluation Tasks}. Our goal is to learn fine-grained, comprehensive, and generalizable representations for e-commerce products, enabling robust performance across diverse downstream tasks. We therefore conduct evaluations on multiple benchmarks, including \bench, M5Product~\cite{dong2022m5product}, and Fashion200K~\cite{han2017automatic}, covering multimodal retrieval, product classification, and attribute prediction. To better demonstrate the quality and generality of the learned representations, 
most baselines
are performed in a zero-shot setting without fine-tuning on the target test distributions. For retrieval tasks, we adopt Recall@$k$ to measure whether the ground-truth candidate appears within the top-$k$ results. For classification and attribute prediction, we use standard metrics, i.e., accuracy, precision, recall, and F1 score. All evaluation protocols are applied across both our method and all baselines to ensure fair comparisons.

\textbf{Baselines}. To evaluate the product understanding capability of \model, we compare it with a range of multimodal representation methods. We first include a large-scale contrastive learning model SigLIP2~\cite{tschannen2025siglip} and recent general multimodal embedding models: Qwen3-VL-Embedding-2B~\cite{li2026qwen3}, GME-Qwen2VL-2B~\cite{zhang2024gme}, and MM-Embed~\cite{lin2024mm}. To further compare with open-source MLLMs on product understanding tasks, we also include InternVL3.5-2B~\cite{wang2025internvl3} and Qwen3-VL-2B~\cite{li2026qwen3}. Since these two models are pretrained with next-token prediction objectives, in addition to zero-shot evaluation, we fine-tune them on our training set and report the results. Finally, we consider domain-specific models for e-commerce, including the dual-flow model FashionCLIP~\cite{chia2022contrastive} and generative-model-based methods CASLIE-S~\cite{ling2024captions}, MOON~\cite{zhang2025moon}, and MOON2.0~\cite{nie2025moon2}.

\subsection{Experimental Results}
Fig.~\ref{fig:radar} shows the performance of \model and the main baselines across various e-commerce tasks. Compared with other baselines, our \model achieves state-of-the-art performance, demonstrating the effectiveness of its reasoning-aware multimodal representation generation paradigm and model architecture in various e-commerce scenarios. Detailed comparisons for each task are provided below, where in all the tables we mark values ranking the first (\textbf{v}) and second (\underline{v}) in each column. The arrows ($\to$) indicate retrieval from query ($q$) to candidate ($c$) across image ($i$), text ($t$), or multimodal ($\text{mm}$) inputs and outputs.

\begin{table}[!t]
    \caption{Ablation study on our \bench benchmark.}
    \label{tab:ablation}
    \centering
    \setlength\tabcolsep{3pt}
    \renewcommand{\arraystretch}{1.0}  
\begin{adjustbox}{width=\linewidth,center}
\begin{tabular}{lccccccc}
\toprule
\multicolumn{1}{c}{\multirow{2}{*}{\diagbox{Ablation}{Metrics}}} & 
\multicolumn{5}{c}{Multimodal Retrieval (R@10)}& 
\multicolumn{1}{c}{Class.}&
\multicolumn{1}{c}{Attr.}\\
\cmidrule(lr){2-6} \cmidrule(lr){7-7} \cmidrule(lr){8-8}

&$q^{i} \to c^{\text{mm}}$
&$q^{t} \to c^{\text{mm}}$
&$q^{\text{mm}} \to c^{\text{mm}}$
&$q^{i} \to c^{t}$
&$q^{t} \to c^{i}$
&Acc.
&Acc.\\

\midrule
full model
& \one{56.39} & \one{48.10} & \one{60.30} & \one{42.95} & \one{38.93} & \one{86.40} & \one{49.92} \\
$w/o$ Reasoning
& 45.26	& 38.33	& 55.09	& 17.12	& 16.77	& 57.52	& 34.21 \\
$w/o$ Modality Fusion
& 51.37	& 41.26	& 56.17	& 35.55	& 31.95	& 82.33	& 43.90 \\
$w/o$ GRPO
& 54.01	& 43.96	& 57.58	& 38.46	& 33.40	& 83.40	& 41.54 \\
$w/o$ \residual
& 52.81	& 43.00	& 57.08	& 37.34	& 32.90	& 80.33	& 42.22 \\

\bottomrule
\end{tabular}
\end{adjustbox}
\end{table}

The results on \bench for multimodal retrieval are summarized in Tab.~\ref{tab:ourbench}. Across all five retrieval tasks, \model consistently achieves the best performance under different values of $k$. Compared with e-commerce models including MOON2.0, CASLIE-S, and FashionCLIP, \model shows substantial improvements, especially in 
cross-modal retrieval tasks. These gains primarily stem from the multi-head modality fusion, which captures complementary cross-modal relationships between images and text across multiple semantic subspaces, thereby enhancing cross-modal alignment and fine-grained feature modeling. Moreover, although the fine-tuned generative baselines, InternVL3.5-2B and Qwen3-VL-2B, improve performance over their zero-shot settings, they remain inferior to \model due to the lack of tailored strategies for e-commerce representation learning. Notably, our method achieves the highest recall with only a $256$-dimensional embedding, which is crucial for real-world applications requiring low latency.

\begin{figure}[t]
  \centering
  \includegraphics[width=\linewidth]{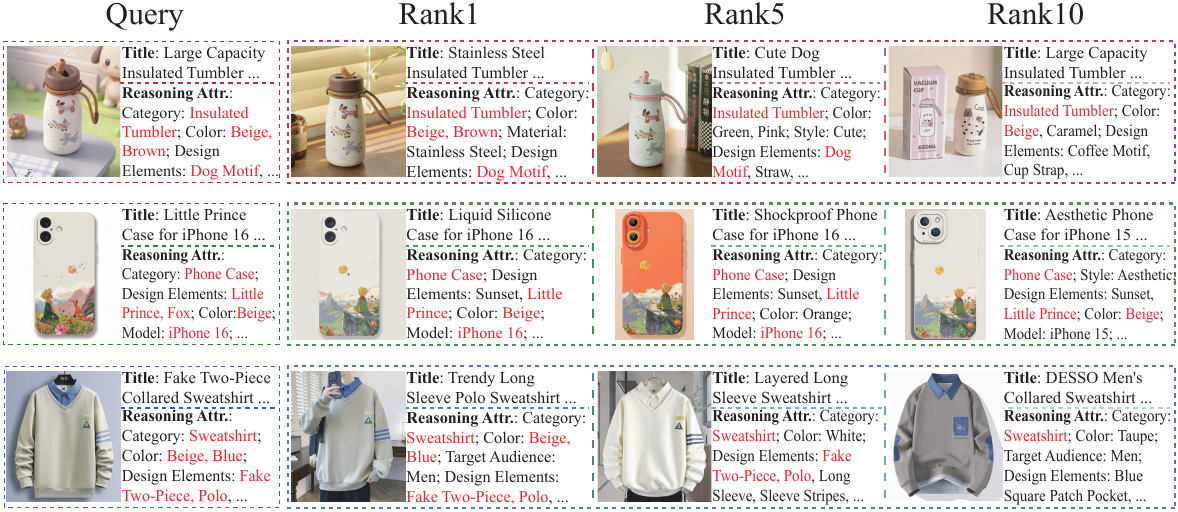}
  \caption{Visualization of retrieval results by our \model for multimodal query-based item retrieval ($q^{\text{mm}}\to c^{\text{mm}}$), with the model-inferred intermediate attributes (``Reasoning Attr.'') displayed alongside each query and its retrieved items.}
  \label{fig:visualizations}
\end{figure}

On product classification and attribute prediction tasks, \model also consistently outperforms all baselines, mainly benefiting from the reasoning paradigm, which explicitly deconstructs multimodal inputs into structured attributes before embedding. Notably, in the most challenging attribute prediction task, it achieves substantial gains over strong e-commerce models such as MOON2.0, demonstrating its ability to preserve and distinguish fine-grained product attributes. This advantage further stems from the joint contrastive and reinforcement learning framework that enables more effective reasoning strategy exploration beyond SFT, and the \residual module, which reinforces the modeling of local details such as color, material, and design, together enabling embeddings that retain the most critical fine-grained information in e-commerce scenarios.


To validate generalization beyond \bench, we evaluate \model on two public benchmarks, M5Product and Fashion200K, as shown in Tab.~\ref{tab:m5product}. Consistent improvements across multimodal retrieval and classification tasks demonstrate the robustness of reasoning-aware representation learning across diverse e-commerce datasets.



\subsection{Ablation Study}
We conduct a systematic ablation study on \bench to quantify the contribution of each component, as shown in Tab.~\ref{tab:ablation}. Ablating any module causes consistent performance degradation across all e-commerce tasks, demonstrating the effectiveness of our integrated design. Notably, removing the reasoning module triggers the most severe degradation (e.g., classification accuracy drops by 28.88\% and $q^{i} \to c^{t}$ retrieval falls by 25.83\%), underscoring its pivotal role in improving representation quality. Removing the multi-head modality fusion primarily attenuates cross-modal retrieval (e.g., -6.98\% in $q^{t} \to c^{i}$), confirming its efficacy in aligning heterogeneous inputs and stabilizing fine-grained correspondences. Similarly, disabling GRPO notably impairs attribute prediction (-8.38\%), demonstrating that the joint contrastive and reinforcement learning framework strengthens fine-grained attribute discrimination. Finally, excluding the \residual module yields a uniform performance decline across all metrics (e.g., -6.07\% in classification and -7.70\% in attribute prediction), verifying its role in reinforcing stable feature propagation.

\subsection{Visualization of Retrieval Results}
As illustrated in Fig.~\ref{fig:visualizations}, we qualitatively evaluate \model on multimodal query-based item retrieval using representations generated by our reasoning-aware framework. We investigate two key aspects by visualizing intermediate reasoning attributes: (1) the model's capability to accurately capture key attributes from image-text inputs, and (2) the positive correlation between the top-ranked retrieval results and the matching consistency of model-inferred attributes. Specifically, the figure presents multimodal queries alongside their top-$k$ retrieved items ($k$ = 1, 5, 10) and corresponding reasoning attributes. For example, in the first row, beyond merely generating global representations, our model further decomposes both the query and candidate items into fine-grained attributes, e.g., ``Category: Insulated Tumbler'', ``Color: Beige, Brown'', and ``Design Elements: Dog Motif''. A comparison of the retrieval results reveals that the Rank~1 item aligns most closely with the query regarding color composition and design elements, whereas lower-ranked items progressively deviate across specific attributes like color and decorative patterns. Ultimately, this attribute-level reasoning empowers the model to derive representations that are highly interpretable and discriminative.

\section{Conclusion}
Breaking from the conventional representation generation paradigm, we propose \model, a reasoning-aware MLLM-based model for e-commerce product understanding that explicitly deconstructs attributes before embedding generation. Our approach incorporates a multi-head modality fusion module, a joint contrastive and reinforcement learning framework, and the \residual module to alleviate attention dilution over long sequences, the imitation bottleneck of SFT, and the attenuation of fine-grained details, respectively. Moreover, we release a large-scale multimodal benchmark \bench covering multiple downstream tasks to support future research on e-commerce product understanding. Overall, our work paves a new avenue for reasoning-aware product representation learning.

\bibliographystyle{ACM-Reference-Format}
\bibliography{samples/sample-base}

@article{yan2025mim,
  title={MIM: Multi-modal Content Interest Modeling Paradigm for User Behavior Modeling},
  author={Yan, Bencheng and Chen, Si and Jia, Shichang and Liu, Jianyu and Liu, Yueran and Fu, Chenghan and Guan, Wanxian and Zhao, Hui and Zhang, Xiang and Zhang, Kai and others},
  journal={arXiv preprint arXiv:2502.00321},
  year={2025}
}

@inproceedings{liu2022pretraining,
  title={Pretraining Representations of Multi-modal Multi-query E-commerce Search},
  author={Liu, Xinyi and Guan, Wanxian and Li, Lianyun and Li, Hui and Lin, Chen and Li, Xubin and Chen, Si and Xu, Jian and Deng, Hongbo and Zheng, Bo},
  booktitle={Proceedings of the 28th ACM SIGKDD Conference on Knowledge Discovery and Data Mining},
  pages={3429--3437},
  year={2022}
}

@inproceedings{li2020adversarial,
  title={Adversarial multimodal representation learning for click-through rate prediction},
  author={Li, Xiang and Wang, Chao and Tan, Jiwei and Zeng, Xiaoyi and Ou, Dan and Ou, Dan and Zheng, Bo},
  booktitle={Proceedings of The Web Conference 2020},
  pages={827--836},
  year={2020}
}

@inproceedings{hendriksen2022multimodal,
  title={Multimodal retrieval in e-commerce: From categories to images, text, and back},
  author={Hendriksen, Mariya},
  booktitle={European Conference on Information Retrieval},
  pages={505--512},
  year={2022},
  organization={Springer}
}

@inproceedings{xu2019open,
  title={Open-world learning and application to product classification},
  author={Xu, Hu and Liu, Bing and Shu, Lei and Yu, P},
  booktitle={The World Wide Web Conference},
  pages={3413--3419},
  year={2019}
}

@inproceedings{dong2022m5product,
  title={M5product: Self-harmonized contrastive learning for e-commercial multi-modal pretraining},
  author={Dong, Xiao and Zhan, Xunlin and Wu, Yangxin and Wei, Yunchao and Kampffmeyer, Michael C and Wei, Xiaoyong and Lu, Minlong and Wang, Yaowei and Liang, Xiaodan},
  booktitle={Proceedings of the IEEE/CVF Conference on Computer Vision and Pattern Recognition},
  pages={21252--21262},
  year={2022}
}

@article{pawlowski2022machine,
  title={Machine learning based product classification for ecommerce},
  author={Paw{\l}owski, Mieczys{\l}aw},
  journal={Journal of Computer Information Systems},
  volume={62},
  number={4},
  pages={730--739},
  year={2022},
  publisher={Taylor \& Francis}
}

@article{yin2024survey,
  title={A survey on multimodal large language models},
  author={Yin, Shukang and Fu, Chaoyou and Zhao, Sirui and Li, Ke and Sun, Xing and Xu, Tong and Chen, Enhong},
  journal={National Science Review},
  volume={11},
  number={12},
  pages={nwae403},
  year={2024},
  publisher={Oxford University Press}
}

@article{tschannen2025siglip,
  title={Siglip 2: Multilingual vision-language encoders with improved semantic understanding, localization, and dense features},
  author={Tschannen, Michael and Gritsenko, Alexey and Wang, Xiao and Naeem, Muhammad Ferjad and Alabdulmohsin, Ibrahim and Parthasarathy, Nikhil and Evans, Talfan and Beyer, Lucas and Xia, Ye and Mustafa, Basil and others},
  journal={arXiv preprint arXiv:2502.14786},
  year={2025}
}

@article{lin2024mm,
  title={Mm-embed: Universal multimodal retrieval with multimodal llms},
  author={Lin, Sheng-Chieh and Lee, Chankyu and Shoeybi, Mohammad and Lin, Jimmy and Catanzaro, Bryan and Ping, Wei},
  journal={arXiv preprint arXiv:2411.02571},
  year={2024}
}

@article{zhang2024gme,
  title={GME: Improving Universal Multimodal Retrieval by Multimodal LLMs},
  author={Zhang, Xin and Zhang, Yanzhao and Xie, Wen and Li, Mingxin and Dai, Ziqi and Long, Dingkun and Xie, Pengjun and Zhang, Meishan and Li, Wenjie and Zhang, Min},
  journal={arXiv preprint arXiv:2412.16855},
  year={2024}
}

@article{liang2025uniecs,
  title={UniECS: Unified Multimodal E-Commerce Search Framework with Gated Cross-modal Fusion},
  author={Liang, Zihan and Ma, Yufei and Qian, ZhiPeng and Dai, Huangyu and Wang, Zihan and Chen, Ben and Lei, Chenyi and Ding, Yuqing and Li, Han},
  journal={arXiv preprint arXiv:2508.13843},
  year={2025}
}

@article{zhang2025moon,
  title={MOON: Generative MLLM-based Multimodal Representation Learning for E-commerce Product Understanding},
  author={Zhang, Daoze and Nie, Zhanheng and Liu, Jianyu and Fu, Chenghan and Guan, Wanxian and Gao, Yuan and Song, Jun and Wang, Pengjie and Xu, Jian and Zheng, Bo},
  journal={arXiv preprint arXiv:2508.11999},
  year={2025}
}

@article{achiam2023gpt,
  title={Gpt-4 technical report},
  author={Achiam, Josh and Adler, Steven and Agarwal, Sandhini and Ahmad, Lama and Akkaya, Ilge and Aleman, Florencia Leoni and Almeida, Diogo and Altenschmidt, Janko and Altman, Sam and Anadkat, Shyamal and others},
  journal={arXiv preprint arXiv:2303.08774},
  year={2023}
}

@article{team2023gemini,
  title={Gemini: a family of highly capable multimodal models},
  author={Team, Gemini and Anil, Rohan and Borgeaud, Sebastian and Alayrac, Jean-Baptiste and Yu, Jiahui and Soricut, Radu and Schalkwyk, Johan and Dai, Andrew M and Hauth, Anja and Millican, Katie and others},
  journal={arXiv preprint arXiv:2312.11805},
  year={2023}
}

@article{bai2023qwen,
  title={Qwen-vl: A frontier large vision-language model with versatile abilities},
  author={Bai, Jinze and Bai, Shuai and Yang, Shusheng and Wang, Shijie and Tan, Sinan and Wang, Peng and Lin, Junyang and Zhou, Chang and Zhou, Jingren},
  journal={arXiv preprint arXiv:2308.12966},
  volume={1},
  number={2},
  pages={3},
  year={2023}
}

@article{chia2022contrastive,
  title={Contrastive language and vision learning of general fashion concepts},
  author={Chia, Patrick John and Attanasio, Giuseppe and Bianchi, Federico and Terragni, Silvia and Magalhaes, Ana Rita and Goncalves, Diogo and Greco, Ciro and Tagliabue, Jacopo},
  journal={Scientific Reports},
  volume={12},
  number={1},
  pages={18958},
  year={2022},
  publisher={Nature Publishing Group UK London}
}

@article{chen2022product2vec,
  title={Product2Vec: Leveraging representation learning to model consumer product choice in large assortments},
  author={Chen, Fanglin and Liu, Xiao and Proserpio, Davide and Troncoso, Isamar},
  journal={NYU Stern School of Business},
  year={2022}
}

@inproceedings{han2017automatic,
  title={Automatic spatially-aware fashion concept discovery},
  author={Han, Xintong and Wu, Zuxuan and Huang, Phoenix X and Zhang, Xiao and Zhu, Menglong and Li, Yuan and Zhao, Yang and Davis, Larry S},
  booktitle={Proceedings of the IEEE international conference on computer vision},
  pages={1463--1471},
  year={2017}
}

@article{ling2025ecommmmu,
  title={EcomMMMU: Strategic Utilization of Visuals for Robust Multimodal E-Commerce Models},
  author={Ling, Xinyi and Du, Hanwen and Zhu, Zhihui and Ning, Xia},
  journal={arXiv preprint arXiv:2508.15721},
  year={2025}
}

@article{jiang2024vlm2vec,
  title={Vlm2vec: Training vision-language models for massive multimodal embedding tasks},
  author={Jiang, Ziyan and Meng, Rui and Yang, Xinyi and Yavuz, Semih and Zhou, Yingbo and Chen, Wenhu},
  journal={arXiv preprint arXiv:2410.05160},
  year={2024}
}

@inproceedings{gao2020fashionbert,
  title={Fashionbert: Text and image matching with adaptive loss for cross-modal retrieval},
  author={Gao, Dehong and Jin, Linbo and Chen, Ben and Qiu, Minghui and Li, Peng and Wei, Yi and Hu, Yi and Wang, Hao},
  booktitle={Proceedings of the 43rd International ACM SIGIR Conference on Research and Development in Information Retrieval},
  pages={2251--2260},
  year={2020}
}

@inproceedings{yu2022commercemm,
  title={Commercemm: Large-scale commerce multimodal representation learning with omni retrieval},
  author={Yu, Licheng and Chen, Jun and Sinha, Animesh and Wang, Mengjiao and Chen, Yu and Berg, Tamara L and Zhang, Ning},
  booktitle={Proceedings of the 28th ACM SIGKDD conference on knowledge discovery and data mining},
  pages={4433--4442},
  year={2022}
}

@inproceedings{dai2024uniembedding,
  title={UniEmbedding: Learning Universal Multi-Modal Multi-Domain Item Embeddings via User-View Contrastive Learning},
  author={Dai, Boqi and Du, Zhaocheng and Zhu, Jieming and Xu, Jintao and Zou, Deqing and Dai, Quanyu and Dong, Zhenhua and Zhang, Rui and Zheng, Hai-Tao},
  booktitle={Proceedings of the 33rd ACM International Conference on Information and Knowledge Management},
  pages={4446--4453},
  year={2024}
}

@article{wang2025internvl3,
  title={Internvl3. 5: Advancing open-source multimodal models in versatility, reasoning, and efficiency},
  author={Wang, Weiyun and Gao, Zhangwei and Gu, Lixin and Pu, Hengjun and Cui, Long and Wei, Xingguang and Liu, Zhaoyang and Jing, Linglin and Ye, Shenglong and Shao, Jie and others},
  journal={arXiv preprint arXiv:2508.18265},
  year={2025}
}

@article{ling2024captions,
  title={Captions Speak Louder than Images (CASLIE): Generalizing Foundation Models for E-commerce from High-quality Multimodal Instruction Data},
  author={Ling, Xinyi and Peng, Bo and Du, Hanwen and Zhu, Zhihui and Ning, Xia},
  journal={arXiv preprint arXiv:2410.17337},
  year={2024}
}

@inproceedings{peng2024ecellm,
  title={eCeLLM: generalizing large language models for E-commerce from large-scale, high-quality instruction data},
  author={Peng, Bo and Ling, Xinyi and Chen, Ziru and Sun, Huan and Ning, Xia},
  booktitle={Proceedings of the 41st International Conference on Machine Learning},
  pages={40215--40257},
  year={2024}
}

@inproceedings{wang2023missrec,
  title={Missrec: Pre-training and transferring multi-modal interest-aware sequence representation for recommendation},
  author={Wang, Jinpeng and Zeng, Ziyun and Wang, Yunxiao and Wang, Yuting and Lu, Xingyu and Li, Tianxiang and Yuan, Jun and Zhang, Rui and Zheng, Hai-Tao and Xia, Shu-Tao},
  booktitle={Proceedings of the 31st ACM International Conference on Multimedia},
  pages={6548--6557},
  year={2023}
}

@article{jiang2024mrse,
  title={MRSE: An Efficient Multi-modality Retrieval System for Large Scale E-commerce},
  author={Jiang, Hao and Zhang, Haoxiang and Hou, Qingshan and Chen, Chaofeng and Lin, Weisi and Zhang, Jingchang and Wang, Annan},
  journal={arXiv preprint arXiv:2408.14968},
  year={2024}
}

@inproceedings{zheng2023delving,
  title={Delving into e-commerce product retrieval with vision-language pre-training},
  author={Zheng, Xiaoyang and Lv, Fuyu and Wang, Zilong and Liu, Qingwen and Zeng, Xiaoyi},
  booktitle={Proceedings of the 46th International ACM SIGIR Conference on Research and Development in Information Retrieval},
  pages={3385--3389},
  year={2023}
}

@inproceedings{li2021embedding,
  title={Embedding-based product retrieval in taobao search},
  author={Li, Sen and Lv, Fuyu and Jin, Taiwei and Lin, Guli and Yang, Keping and Zeng, Xiaoyi and Wu, Xiao-Ming and Ma, Qianli},
  booktitle={Proceedings of the 27th ACM SIGKDD Conference on Knowledge Discovery \& Data Mining},
  pages={3181--3189},
  year={2021}
}

@inproceedings{jin2023learning,
  title={Learning instance-level representation for large-scale multi-modal pretraining in e-commerce},
  author={Jin, Yang and Li, Yongzhi and Yuan, Zehuan and Mu, Yadong},
  booktitle={Proceedings of the IEEE/CVF Conference on Computer Vision and Pattern Recognition},
  pages={11060--11069},
  year={2023}
}

@article{zhu2025internvl3,
  title={Internvl3: Exploring advanced training and test-time recipes for open-source multimodal models},
  author={Zhu, Jinguo and Wang, Weiyun and Chen, Zhe and Liu, Zhaoyang and Ye, Shenglong and Gu, Lixin and Tian, Hao and Duan, Yuchen and Su, Weijie and Shao, Jie and others},
  journal={arXiv preprint arXiv:2504.10479},
  year={2025}
}

@article{fu2025moon,
  title={MOON Embedding: Multimodal Representation Learning for
E-commerce Search Advertising},
  author={Fu, Chenghan and Zhang, Daoze and Lin, Yukang and Nie, Zhanheng and Zhang, Xiang and Liu, Jianyu and Liu, Yueran and Guan, Wanxian and Wang, Pengjie and Xu, Jian and Zheng, Bo},
  journal={arXiv preprint},
  year={2025}
}

@article{nie2025moon2,
  title={MOON2. 0: Dynamic Modality-balanced Multimodal Representation Learning for E-commerce Product Understanding},
  author={Nie, Zhanheng and Fu, Chenghan and Zhang, Daoze and Wu, Junxian and Guan, Wanxian and Wang, Pengjie and Xu, Jian and Zheng, Bo},
  journal={arXiv preprint arXiv:2511.12449},
  year={2025}
}

@article{cui2025think,
  title={Think then embed: Generative context improves multimodal embedding},
  author={Cui, Xuanming and Cheng, Jianpeng and Chen, Hong-you and Shukla, Satya Narayan and Awasthi, Abhijeet and Pan, Xichen and Ahuja, Chaitanya and Mishra, Shlok Kumar and Yang, Yonghuan and Xiao, Jun and others},
  journal={arXiv preprint arXiv:2510.05014},
  year={2025}
}

@article{bai2025qwen3,
  title={Qwen3-vl technical report},
  author={Bai, Shuai and Cai, Yuxuan and Chen, Ruizhe and Chen, Keqin and Chen, Xionghui and Cheng, Zesen and Deng, Lianghao and Ding, Wei and Gao, Chang and Ge, Chunjiang and others},
  journal={arXiv preprint arXiv:2511.21631},
  year={2025}
}

@article{meng2024deepstack,
  title={Deepstack: Deeply stacking visual tokens is surprisingly simple and effective for lmms},
  author={Meng, Lingchen and Yang, Jianwei and Tian, Rui and Dai, Xiyang and Wu, Zuxuan and Gao, Jianfeng and Jiang, Yu-Gang},
  journal={Advances in Neural Information Processing Systems},
  volume={37},
  pages={23464--23487},
  year={2024}
}

@inproceedings{zhu2020multimodal,
  title={Multimodal joint attribute prediction and value extraction for e-commerce product},
  author={Zhu, Tiangang and Wang, Yue and Li, Haoran and Wu, Youzheng and He, Xiaodong and Zhou, Bowen},
  booktitle={Proceedings of the 2020 Conference on Empirical Methods in Natural Language Processing (EMNLP)},
  pages={2129--2139},
  year={2020}
}

@article{jiang2024e5,
  title={E5-v: Universal embeddings with multimodal large language models},
  author={Jiang, Ting and Song, Minghui and Zhang, Zihan and Huang, Haizhen and Deng, Weiwei and Sun, Feng and Zhang, Qi and Wang, Deqing and Zhuang, Fuzhen},
  journal={arXiv preprint arXiv:2407.12580},
  year={2024}
}

@article{hu2026adaptive,
  title={Adaptive Global and Fine-Grained Perceptual Fusion for MLLM Embeddings Compatible with Hard Negative Amplification},
  author={Hu, Lexiang and Xue, Youze and Li, Dian and Liu, Gang and Lin, Zhouchen},
  journal={arXiv preprint arXiv:2602.05729},
  year={2026}
}

@article{tang2025large,
  title={Large Reasoning Embedding Models: Towards Next-Generation Dense Retrieval Paradigm},
  author={Tang, Jianting and Li, Dongshuai and Wen, Tao and Lv, Fuyu and Ou, Dan and Xu, Linli},
  journal={arXiv preprint arXiv:2510.14321},
  year={2025}
}

@article{wei2022chain,
  title={Chain-of-thought prompting elicits reasoning in large language models},
  author={Wei, Jason and Wang, Xuezhi and Schuurmans, Dale and Bosma, Maarten and Xia, Fei and Chi, Ed and Le, Quoc V and Zhou, Denny and others},
  journal={Advances in neural information processing systems},
  volume={35},
  pages={24824--24837},
  year={2022}
}

@article{wu2025language,
  title={When language overrules: Revealing text dominance in multimodal large language models},
  author={Wu, Huyu and Tang, Meng and Zheng, Xinhan and Jiang, Haiyun},
  journal={arXiv preprint arXiv:2508.10552},
  year={2025}
}

@inproceedings{qin2022devil,
  title={The devil in linear transformer},
  author={Qin, Zhen and Han, Xiaodong and Sun, Weixuan and Li, Dongxu and Kong, Lingpeng and Barnes, Nick and Zhong, Yiran},
  booktitle={Proceedings of the 2022 Conference on Empirical Methods in Natural Language Processing},
  pages={7025--7041},
  year={2022}
}

@article{xiao2023efficient,
  title={Efficient streaming language models with attention sinks},
  author={Xiao, Guangxuan and Tian, Yuandong and Chen, Beidi and Han, Song and Lewis, Mike},
  journal={arXiv preprint arXiv:2309.17453},
  year={2023}
}

@article{liu2025uft,
  title={Uft: Unifying supervised and reinforcement fine-tuning},
  author={Liu, Mingyang and Farina, Gabriele and Ozdaglar, Asuman},
  journal={arXiv preprint arXiv:2505.16984},
  year={2025}
}

@article{wang2026learning,
  title={Learning While Staying Curious: Entropy-Preserving Supervised Fine-Tuning via Adaptive Self-Distillation for Large Reasoning Models},
  author={Wang, Hao and Gu, Hao and Piao, Hongming and Gong, Kaixiong and Ye, Yuxiao and Yue, Xiangyu and Han, Sirui and Guo, Yike and Wu, Dapeng},
  journal={arXiv preprint arXiv:2602.02244},
  year={2026}
}

@inproceedings{gu2025breaking,
  title={Breaking the modality barrier: Universal embedding learning with multimodal llms},
  author={Gu, Tiancheng and Yang, Kaicheng and Feng, Ziyong and Wang, Xingjun and Zhang, Yanzhao and Long, Dingkun and Chen, Yingda and Cai, Weidong and Deng, Jiankang},
  booktitle={Proceedings of the 33rd ACM International Conference on Multimedia},
  pages={2860--2869},
  year={2025}
}

@inproceedings{chu2024navigate,
  title={Navigate through enigmatic labyrinth a survey of chain of thought reasoning: Advances, frontiers and future},
  author={Chu, Zheng and Chen, Jingchang and Chen, Qianglong and Yu, Weijiang and He, Tao and Wang, Haotian and Peng, Weihua and Liu, Ming and Qin, Bing and Liu, Ting},
  booktitle={Proceedings of the 62nd Annual Meeting of the Association for Computational Linguistics (Volume 1: Long Papers)},
  pages={1173--1203},
  year={2024}
}

@inproceedings{lai2024lisa,
  title={Lisa: Reasoning segmentation via large language model},
  author={Lai, Xin and Tian, Zhuotao and Chen, Yukang and Li, Yanwei and Yuan, Yuhui and Liu, Shu and Jia, Jiaya},
  booktitle={Proceedings of the IEEE/CVF conference on computer vision and pattern recognition},
  pages={9579--9589},
  year={2024}
}

@inproceedings{xu2025llava,
  title={Llava-cot: Let vision language models reason step-by-step},
  author={Xu, Guowei and Jin, Peng and Wu, Ziang and Li, Hao and Song, Yibing and Sun, Lichao and Yuan, Li},
  booktitle={Proceedings of the IEEE/CVF International Conference on Computer Vision},
  pages={2087--2098},
  year={2025}
}

@article{yan2025o1,
  title={O1 embedder: Let retrievers think before action},
  author={Yan, Ruiran and Liu, Zheng and Lian, Defu},
  journal={arXiv preprint arXiv:2502.07555},
  year={2025}
}

@article{cui2025reason,
  title={Reason to Contrast: A Cascaded Multimodal Retrieval Framework},
  author={Cui, Xuanming and Chen, Hong-You and Yu, Hao and Yuan, Hao and Wang, Zihao and Mishra, Shlok Kumar and Yu, Hanchao and Yang, Yonghuan and Xiao, Jun and Lim, Ser-Nam and others},
  journal={arXiv preprint arXiv:2602.23369},
  year={2025}
}

@article{hao2026trace,
  title={TRACE: Task-Adaptive Reasoning and Representation Learning for Universal Multimodal Retrieval},
  author={Hao, Xiangzhao and Wang, Shijie and Yang, Tianyu and Wang, Tianyue and Guo, Haiyun and Wang, JinQiao},
  journal={arXiv preprint arXiv:2603.02929},
  year={2026}
}

@article{jiang2026embed,
  title={Embed-RL: Reinforcement Learning for Reasoning-Driven Multimodal Embeddings},
  author={Jiang, Haonan and Wang, Yuji and Zhu, Yongjie and Lu, Xin and Qin, Wenyu and Wang, Meng and Wan, Pengfei and Tang, Yansong},
  journal={arXiv preprint arXiv:2602.13823},
  year={2026}
}

@inproceedings{zhang2025enhancing,
  title={Enhancing chain of thought prompting in large language models via reasoning patterns},
  author={Zhang, Yufeng and Wang, Xuepeng and Wu, Lingxiang and Wang, Jinqiao},
  booktitle={Proceedings of the AAAI Conference on Artificial Intelligence},
  volume={39},
  number={24},
  pages={25985--25993},
  year={2025}
}

@article{li2026qwen3,
  title={Qwen3-VL-Embedding and Qwen3-VL-Reranker: A Unified Framework for State-of-the-Art Multimodal Retrieval and Ranking},
  author={Li, Mingxin and Zhang, Yanzhao and Long, Dingkun and Chen, Keqin and Song, Sibo and Bai, Shuai and Yang, Zhibo and Xie, Pengjun and Yang, An and Liu, Dayiheng and others},
  journal={arXiv preprint arXiv:2601.04720},
  year={2026}
}

@article{vaswani2017attention,
  title={Attention is all you need},
  author={Vaswani, Ashish and Shazeer, Noam and Parmar, Niki and Uszkoreit, Jakob and Jones, Llion and Gomez, Aidan N and Kaiser, {\L}ukasz and Polosukhin, Illia},
  journal={Advances in neural information processing systems},
  volume={30},
  year={2017}
}

@article{chen2025sft,
  title={Sft or rl? an early investigation into training r1-like reasoning large vision-language models},
  author={Chen, Hardy and Tu, Haoqin and Wang, Fali and Liu, Hui and Tang, Xianfeng and Du, Xinya and Zhou, Yuyin and Xie, Cihang},
  journal={arXiv preprint arXiv:2504.11468},
  year={2025}
}

@inproceedings{chu2025sft,
  title={SFT Memorizes, RL Generalizes: A Comparative Study of Foundation Model Post-training},
  author={Chu, Tianzhe and Zhai, Yuexiang and Yang, Jihan and Tong, Shengbang and Xie, Saining and Schuurmans, Dale and Le, Quoc V and Levine, Sergey and Ma, Yi},
  booktitle={International Conference on Machine Learning},
  pages={10818--10838},
  year={2025},
  organization={PMLR}
}










\end{document}